\documentclass[conference]{IEEEtran}
\IEEEoverridecommandlockouts
\usepackage{cite}
\usepackage{amsmath,amssymb,amsfonts}
\usepackage{algorithmic}
\usepackage{graphicx}
\usepackage{textcomp}
\usepackage{makecell}
\usepackage{xcolor}
\usepackage{wrapfig}
\usepackage{listings}

\usepackage{soul}

\usepackage{todonotes}

\begin{document}

\title{Using machine learning to reduce ensembles of geological models for oil and gas exploration}

\author{
\IEEEauthorblockN{Anna Roub\'{i}\v{c}kov\'{a}}
\IEEEauthorblockA{\textit{EPCC, The University of Edinburgh}\\
a.roubickova@epcc.ed.ac.uk}
\\
\IEEEauthorblockN{Lucy MacGregor}
\IEEEauthorblockA{\textit{Cognitive Geology}\\
lucy.macgregor@cognitivegeology.com}

\and 

\IEEEauthorblockN{Nick Brown}
\IEEEauthorblockA{\textit{EPCC, The University of Edinburgh}\\
n.brown@epcc.ed.ac.uk}
\\ 
\\ \\ \\ 
\and

\IEEEauthorblockN{Oliver Thomson Brown}
\IEEEauthorblockA{\textit{EPCC, The University of Edinburgh}\\
o.brown@epcc.ed.ac.uk}
\\
\IEEEauthorblockN{Mike Stewart}
\IEEEauthorblockA{\textit{Cognitive Geology}\\
mike.stewart@cognitivegeology.com}
}


\maketitle

\begin{abstract}
Exploration using borehole drilling is a key activity in determining the most appropriate locations for the petroleum industry to develop oil fields. However, estimating the amount of Oil In Place (OIP) relies on computing with a very significant number of geological models, which, due to the ever increasing capability to capture and refine data, is becoming infeasible. As such, data reduction techniques are required to reduce this set down to a smaller, yet still fully representative ensemble. In this paper we explore different approaches to identifying the key grouping of models, based on their most important features, and then using this information select a reduced set which we can be confident fully represent the overall model space. The result of this work is an approach which enables us to describe the entire state space using only 0.5\% of the models, along with a series of lessons learnt. The techniques that we describe are not only applicable to oil and gas exploration, but also more generally to the HPC community as we are forced to work with reduced data-sets due to the rapid increase in data collection capability.
\end{abstract}



\section{Introduction}
Oil and gas companies rely on information gathered from exploratory boreholes to make decisions around oil field development. Necessarily sparse, borehole observations in the form of well logs \cite{crain2002crain} are fed into computational and analytic models which predict properties of the geology and from this conclusions are drawn. Bearing in mind the very significant cost of further exploration and oilfield development, the industry is constantly looking for improvements to these models and ability to leverage their data more accurately \cite{swoop}.

The current state of the art is to generate tens of millions of possible geological models from the well log data. Whilst geologists would ideally like to work with a full set of models to maintain a full representation, this is simply not feasible because of the computation involved in each model. Given the current technology available to oil and gas companies, and the need for timely results, existing approaches rely on simplification methods \cite{crain2002crain} which significantly reduce the number of models. This reduced ensemble contains far fewer models than the original set, but this is currently  done on a fairly arbitrary basis, and the resulting simplified ensemble can be very sensitive, exhibiting a great deal of variability in associated target properties, such as total amount of hydrocarbons present. Whilst the geologists presently have no choice, the variability around these simplified ensembles can cause high uncertainty of the oilfield's properties. 

The fundamental problem here is that the data reduction approach adopted increases the likelihood that the simplified model ensemble is affected by \emph{black swan events}. This is where the geology exhibits unexpected behaviour, typically due to highly unlikely scenarios taking place, that can not be captured by analysis of the reduced data set. Black swans are a problem because, ultimately, missing these can result in poor decisions being made and significant financial impact or lost opportunities. To guard against these black swans events, it might seem like one would need to go back to using the \emph{full} model set, however this is infeasible given the scale of the problem involved. 

As such, a different approach must be adopted, which is both computationally feasible and less likely to miss anomalous conditions in the geology. It is our believe that one such approach is to be more insightful when selecting the reduced set of models, and to ensure that the reduced ensemble is fully representative of the overall geology. In this work we explore the reduction of models to a manageable level, whilst maintaining an appropriate range of possible values of the target properties, using machine learning techniques. Using these methods we can reduce the effort required to evaluate the target properties of an oilfield for different models, and from this explore the similarity of models with respect to the target properties, effectively serving as a metric for grouping the models based on clustering techniques.

This paper is organised as follows; in Section \ref{sec:bg} we define the problem in more detail and explore some of the related work already done by the petroleum industry in similar areas.
In Section~\ref{sec:modelreduction} we explore possible techniques that can automate ensemble reduction by clustering models together, such that an individual model from each cluster represents the geological properties of the area. We identify the issues related to defining a suitable metric function for the clustering algorithms, and in Section~\ref{sec:ml} explore the role of supervised machine learning in approximating a suitable distance function. Section~\ref{sec:ss-red} combines the findings of the previous two sections, deriving an approach that enables us to reduce the ensemble to around 0.5\% of its original size while maintaining a fully representative set of models. Lastly we discuss conclusions and explore further work in Section~\ref{sec:conc}.



\section{Background and related work}
\label{sec:bg}
Broadly speaking, the goal of the industry is to optimize hydrocarbon production in the most cost effective manner, and for that petroleum engineers need to asses the oilfield as a whole. This assessment is based on observations of the subsurface recorded in \emph{well logs}, where a borehole is drilled into the surface and various probes are lowered to record physical properties, such as neutron porosity, temperature and gamma rays penetration. These measurements are then interpreted by a petrophysicist into higher level properties of the rock surrounding the borehole, such as its porosity, permeability, fluid saturation, and geological makeup.

The location of hydrocarbon reservoirs, which is where the oil and gas can be found, depends on the structure of the subsurface. Even though the well logs are very detailed, on average containing around 20000 rows of measurements for dozens of basic physical and higher level rock properties, they cover the oilfield rather sparsely and this makes it generally hard to predict the structure of the subsurface. To reduce the volume of data, and to aid the exploration of the subsurface, petrophysicists describe the oilfield by means of geological \emph{trends}, and define plausible geological models capturing the functional relationship between these trends and the rock properties. The petroleum industry is mainly concerned with the following four trends:
\begin{itemize}
	\item \textbf{Depth} describes how rock properties, such as density, change with depth.
	\item \textbf{Stratigraphy} describes how rock properties evolve based on geological strata undergoing similar geomorphological processes.
	\item \textbf{Strike} describes how the intersection of the well log with the horizontal plane changes.
	\item \textbf{Dip} describes how the steepest angle of descent relative \\ to the horizontal plane changes.
\end{itemize}

In practice, the evolution of a specific property with respect to a given trend defines a function, and this function can be represented by its knot points. Effectively this means that trends can be represented as a sequence of numbers, and the Depth, Dip, and Strike trends are fairly simple and in this work defined by three knot points each. By contrast the stratigraphic trend is more complex and described using 35 knot points. The concatenation of different trend sequences is called a \emph{gene}, and corresponds to a sequence of 44 numbers in the work described in this paper. 


When defining the geological model of an oilfield, geologists focus on a set of properties of interest and based on these identify all the possible genes that are in line with the logged measurements. In the work described in this paper, we use 24 genes for each geological property and to keep the problem size manageable, we consider only three properties. These are water saturation, rock porosity and net-to-gross, all three of which are independent from each other, and therefore any combination of their genes is entirely possible and valid. The ordered sequence of the properties' genes is referred to as the \emph{genome}, and this defines a geological model of the oilfield, with the model ensemble containing all possible genomes. Based on this, the ultimate metric of interest is the Oil-in-Place (OIP) value, which is the amount of crude oil estimated to be in a reservoir. 

The work described in this paper is based upon a synthetic set of data that describes an \emph{idealised} oilfield. This is a comparable representation of what can be found in an average oilfield and is described by the four trends above, tracking three rock properties. As already mentioned, in our dataset every property has 24 different but equally possible explanations. These are the genes, and they are captured as knot points of the functional evolution of the properties with respect to the various trends, resulting in a 44-dimensional vector. Genes are sequentially numbered, with the number identifying a specific gene called an \emph{allele}.

The genes for different properties are concatenated into an ordered sequence, the genome, and for this oilfield a full ensemble of models describing the 3 properties, each with 24 genes, results in $24^3$ (13824) possible geological models. In reality, more properties are of interest and as such the number of models can easily run into the tens of millions. However, whilst the number and makeup of models explored in this paper represents a simplified problem, being able to highlight similarities and effectively reduce the number in this simplified set is a very significant first step, and the lessons learnt can then apply to larger model sizes. 

Every model in our idealised oilfield can be defined by
\begin{itemize}
	\item The concatenation of the three property genes describing each trend explicitly, resulting in a vector of 132 numbers.
	\item A triplet alleles, that is, a triplet of natural numbers from zero to twenty three encoding which gene describes the evolution of which property.
	\item A sequential identification number, which uniquely identifies that geological model and is based on the alleles used in the model definition.
\end{itemize}
  
For example, a model described by a triplet \emph{[5, 7, 13]} would be allocated \emph{3061} as its identification number ($5 \times 24^2 + 7 \times 24 + 13$ ). The triplet of alleles can be also be considered as a dictionary, where the first property, in this case saturation (sw), is defined by gene number~5, the second property, net-to-gross (ntg), is defined by gene number~7, and the third property, porosity (phi), is defined by gene number~13. Therefore, given
\begin{itemize}
	\item sw[5] = [0.077, 0.024, -0.029, -0.315, 0.087, ..., 0.073]
	\item ntg[7] = [0.621, 0.603, 0.584, 0.126, -0.035, ..., 0.217] 
	\item phi[13] = [-0.041, -0.014, 0.014, -0.013,	0.004, ..., 0.209]
\end{itemize}
then the full genome of this model with identify number 3061, is the concatenation of sw[5], ntg[7] and phi[13], resulting in a vector of \emph{[0.077, 0.024, -0.029, -0.315, ..., 0.209]}.

In \cite{zhang2018efficient}, the authors used a technique called oil reservoir history matching (HM) for estimating oil reservoir models parameters and making production forecasts. Whilst this is a different problem area from the one we are concerned with here, focused on existing oil reservoirs rather than exploration for new reservoirs, they had similar issues in terms of the size of the data. To address this, they used dimensionality reduction, discovering low-dimensional representations of high-dimensional observational data. They adopted a method based on logistic regression, and found that this actually improved the quality of their insights due to the reduction of noise. The data they were working with is very different from our geological models, as our data represent a much larger area and conditions between one model and another can be very different. Nevertheless, the fact that they demonstrated good results using only a subset of their data strengthens our hypothesis that a data reduction approach could work for our problem as well.

As already described, simulating reservoirs at high levels of accuracy is very computationally expensive. An alternative approach is to keep the full set of models but reduce the number of properties in a model instead. This approach was the basis of work done in \cite{he2013reduced}, simulating reservoir properties based on an ensemble of models, each containing a reduced set of properties. They used a Kalman filter \cite{welch1995introduction} and their approach provided results in reasonably close agreement with high fidelity simulations which included all the data. However, the speedups obtained for running with the reduced data set were modest and much of this was due to the overhead of their reduction approach. Encoding a geological model using the genetical approach describe above already reduces the size of the model description considerably in comparison to the full well logs and can not be reduced without significant impact on the the quality of the model's definition.

%
%



\section{Identifying groups of similar models}
\label{sec:modelreduction}


Real-world exploration involves working with many millions of geological models, and it is simply not realistic to explore target properties of each of these models manually. As such, a key question is whether machine learning (ML) techniques can be used to identify groups of models with similar target properties, allowing one to effectively reduce a large number of models into a much smaller but still representative subset. This involves grouping models by their geological behaviours, and is necessary in order to identify the range of distinct scenario outcomes in the dataset. 
\subsection*{Exploratory Analysis}  
\label{ss:histogram}

In this section, we explore the application of unsupervised learning techniques to identify groups of models according to a similarity metric, without knowing the correct grouping labels beforehand. Existing computational approaches can calculate the OIP, and then use this to label the dataset for use with machine learning training and testing. However, grouping models into a number of distinct clusters based upon the similarity of model properties is far more subtle. This is because the geologists do not have a clear set of rules to determine whether one model fits into one group or another, meaning that any manual labelling for supervised learning would be highly unreliable. Instead, using unsupervised techniques to identify similarities that might not be obvious to the human, is desirable.

The first step in clustering the models is to decide what information should be used to drive such a grouping, which involves identifying what information describes the behaviour of geological models sufficiently well. As described in Section \ref{sec:bg}, the model's identity number, tuple of alleles identifying property genes, and the model's genome, provides a complete definition of the model. Whilst each of these properties are interrelated, the key question is which is most suited to represent the individual nature of a specific model and the target OIP.

\begin{figure}[htb]
\centering
\includegraphics[scale=.5]{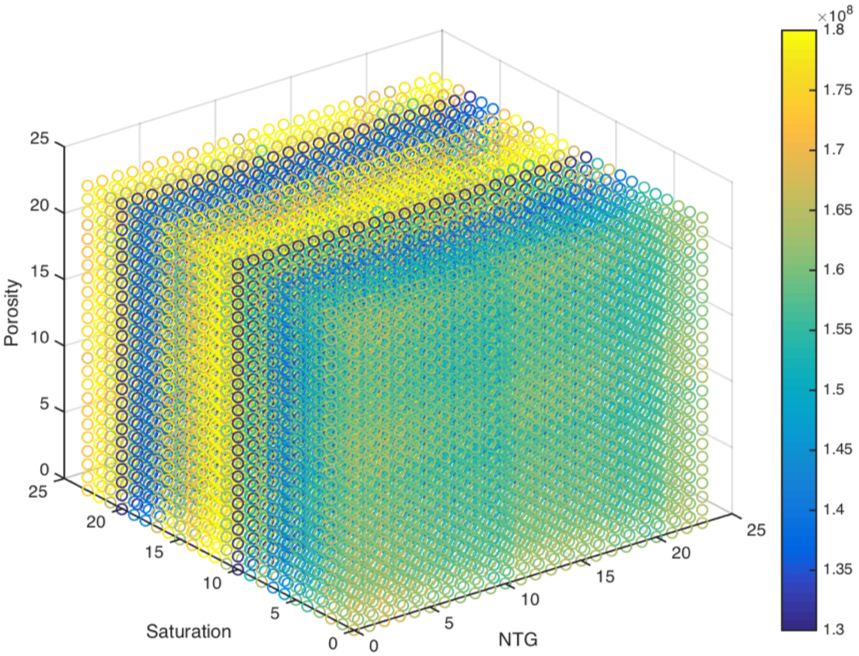}
\caption{The space defined by property alleles (indexes). The axes correspond to the three input properties and the natural numbers on them mark different alleles. The geological models correspond to points in this space, where every dimension identifies which gene was used in the model for a given property. The models are colour-coded by their associates target property (OIP).}
\label{fig:genes2oip-a}
\end{figure}

The unique sequential model number on its own is not very useful for the clustering, as it does not carry any explicit information about the model properties. The tuple of alleles is also not particularly useful as this is an index into detailed property information, rather than that information itself. Figure~\ref{fig:genes2oip-a} illustrates how the OIP depends on the properties, and demonstrates how the alleles cover the whole input space uniformly, not providing sufficient discrimination for clustering. 

Therefore the third property, the models' genomes, has been selected as the property to cluster on. In order to reduce the model ensemble, we needed to identify the clusters which coincide with groups of models that exhibit similar geological behaviours, or, in other words, have similar associated OIP. 

\begin{figure}[htb]
\centering
\begin{tabular}{c c}
	\includegraphics[scale=.32]{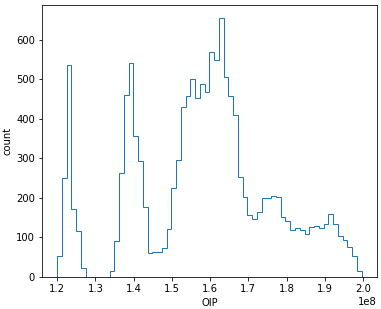} & \includegraphics[scale=.27]{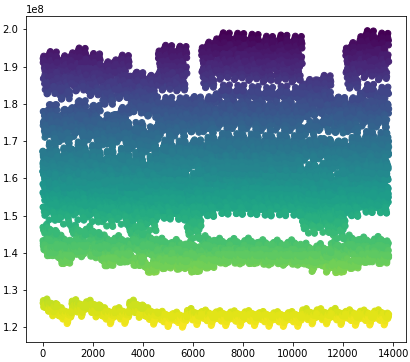} \\
		(a) & (b) \\
\end{tabular}
\caption{(a) An equi-width histogram, where every bin defines an ideal cluster of models with respect to the models' OIP and (b) a visualisation of how the histogram-defined clusters correlate to models' OIP: the $x$-axis corresponds to model identity numbers across the full ensemble, the $y$-axis represents the associated OIP. Points in (b) are colour-coded by the cluster they belong to according to the histogram-based clustering of (a).}
\label{fig:genes2oip-b}
\end{figure}

Figure~\ref{fig:genes2oip-b}~(a) illustrates a histogram of OIP values across 64 bins. These bins span equal ranges of the OIP value, and so every bin can be viewed as a cluster of similarly behaving models. Figure~\ref{fig:genes2oip-b}~(b) visualises these clusters based on the simple binning approach, where the $x$-axis corresponds to model numbers, and the $y$-axis shows the models' OIP. In this plot every observation is colour-coded by the cluster it belongs to and we can see that the clusters define horizontal layers in the plot according to the OIP values.

Figure~\ref{fig:genes2oip-b}~(b) can be considered the gold standard of geological model clustering where, by using these bins, we can group individual models together. The reduced set of ensemble models that cover the entire property space is then formed by selecting individual models from each of the different clusters. However, the problem with this approach is that, in order to build the histogram of bins in the first place, the geologists must know the OIP, and also decide how many bins they are using. Figure~\ref{fig:genes2oip-b}~(b) relies on the OIP label already provided in the data, and we selected 64 bins based on empirical experience. In the real-world, at this point in the workflow, the OIP is unknown because this is only determined later on by evaluating each of the individual models. As such, clustering methods that do not require the OIP and explicit number of clusters to be provided are required. 

In the remainder of this section we focus on density-based clustering and self-organising feature maps. These unsupervised learning techniques have the potential to produce the desired groupings without the explicit OIP figure. By comparing the resulting clusters against the gold standard of Figure~\ref{fig:genes2oip-b}~(b), we are able to determine whether the unsupervised approaches are able to match fidelity, without relying on the OIP value.

\subsection{Density-based Clustering} 
Density-based Clustering \cite{ester1996density} is an unsupervised learning algorithm that defines clusters as groups of observations belonging to the same neighbourhood in the clustering space, with respect to a given similarity metric. This neighbourhood is the minimum number of observations (denoted \emph{min\_samples}) that differentiate a cluster from outliers, and by a maximum distance at which two observations are still considered similar (denoted $\varepsilon$). 

It can be seen in the histogram of Figure~\ref{fig:genes2oip-b}~(a) that the majority of models have a mid-level OIP figure, with far fewer models containing very high or very low OIP. Therefore the desired clusters need to be of different sizes and this was the reason why density-based clustering was chosen. In contrast to other more common clustering algorithms, such as k-means \cite{jain2010data}, density-based clustering avoids imposing any significant constraints on the shapes or sizes of the clusters. In addition, this clustering approach does not require one to provide, or guess, the total number of clusters beforehand. This is important because it allows us to define the desired reduction qualitatively, in terms of the similarity of model properties, rather than quantitatively which would have involved stating how many distinct groups of models we would like the model to produce.

For this work we used the DBSCAN algorithm \cite{schubert2017dbscan} with $\varepsilon$ as 0.6, \emph{min\_samples} as 10, and Euclidean distance as the similarity metric. In this set-up, genomes are interpreted in a purely geometrical way, where each model is treated as a point in $n$-dimensional space. $n$ is the total number of knot-points needed to describe all the property-trends, which is effectively the genome's length. Clusters of models are then formed based on the distance between genomes, but the drawback we found with this approach is that the metric gives every component equal weight. This is not ideal because, as illustrated in Figure~\ref{fig:genes2oip-a}, different properties impact the OIP differently and hence distinct weights based on the impact would be far better.

\begin{figure}[htb]
	\begin{center}
		\includegraphics[width=0.25\textwidth]{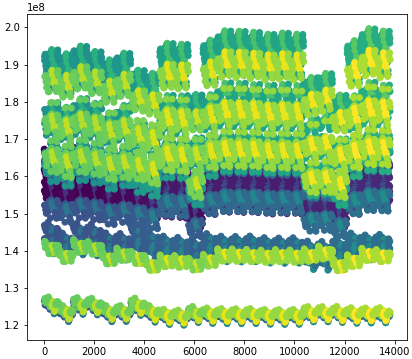}
	\end{center}
	\caption{Groups of geological models identified by DBSCAN using Euclidean distance as the metric. The model identity number across the full ensemble is on the $x$-axis, and the $y$-axis shows the models' OIP. Models are colour-coded by the cluster they belong to.}
	\label{fig:clusters}
\end{figure}

Figure~\ref{fig:clusters} illustrates the resulting model groupings when using density-based clustering. While some of the clusters coincide with similar OIP values, for instance the darker layers where OIP is about $1.55 \times 10^8$, many of the groupings do not. It can be seen in Figure~\ref{fig:clusters} that many clusters cover a wide vertical spread, representing a large range of OIP values. This is undesirable, and requires the use of alternative metrics, replacing Euclidean distance. A different metric would enable us to select different weights for different trends and genes. However, the downside with using an alternative metric in this manner is that it significantly increases the computational complexity of the DBSCAN algorithm. This is because the algorithm needs to either recompute pair-wise distances for all the element quadratic number of times, with time complexity $O(n^2)$ for $n$ models, or store an all-pairs distance matrix of size $n \times n$. In the test set of our study, the number of models is fairly small, $n = 13824$, whereas real-world applications involve model sets which are multiple orders of magnitude larger. This makes the computation infeasible, either due to computation limits of memory limits, and as such we concluded that, for density-based clustering, the Euclidean distance metric is the only one practical for use with real-world geological models.

\subsection{Self-organising Feature Maps}  

Self-organising Feature Maps (SOFM) \cite{foody1999applications} are a form of artificial neural network that uses Kohonen’s learning rule \cite{Demuth:2014:NND:2721661} and a competitive layer to perform unsupervised learning. These networks are commonly used to generate transitional states among sparse data points and to organise the points such that similar elements are nearer each other. Effectively this approach reduces the dimensionality of input data to two dimensions. 



The neurons in the SOFM's hidden layer are typically organised in a two dimensional, rectangular grid. The result is that the network topology is one of the major defining parameters of SOFM, and as such the dimensions of the grid and hence the number of neurons must be selected carefully. Every neuron in the hidden layer is fully connected to the input layer and weights of the connections associated with a neuron can be thought of as coordinates of that individual neuron in a space of appropriate dimensionality.

The competitive behaviour of the SOFM identifies which neuron is  most similar to a presented observation, in our case a model's genome, according to some metric. This neuron's weights are then adjusted so that the neuron becomes even closer to the presented observation, which leads to two further properties that define the behaviour of SOFM. These are the metric function, which dictates the similarity between two elements, and the learning rate, which controls by how much neurons changes their weights. 

The learning step in SOFM usually affects a neighbourhood of the most similar neuron. The algorithm begins by updating all neurons in the SOFM, where the impact of new observations on neuron weights decreases with their increasing distance from the observation. In addition, the size of the neighbourhood reduces as the training progresses, with only one or very few neurons being updated at the end of the training. Consequently, the neurons of the SOFM end up organised such that the ones close to each other react to similar inputs. Therefore, once the map is fitted, we can define a cluster of geological models as the set of geological models that are mapped onto the same neuron.

To provide a direct comparison against the other approaches, we set the number of neurons in the network equal to the number of clusters identified by DBSCAN and that found emperically, which was 64. This results in a network with an \emph{8 by 8} topology, using Euclidean distance as the metric. A cluster of geological models is formed by grouping all the models which map to the same neuron. 

\begin{figure}
	\begin{center}
	\begin{tabular}{c c}
		\includegraphics[scale=.161]{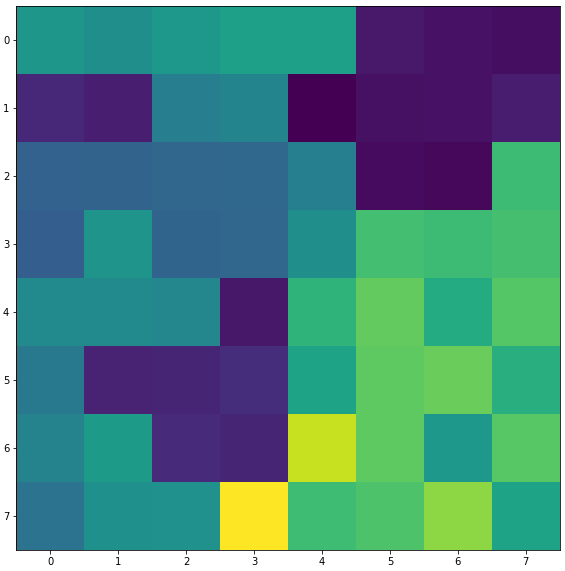} & \includegraphics[scale=.27]{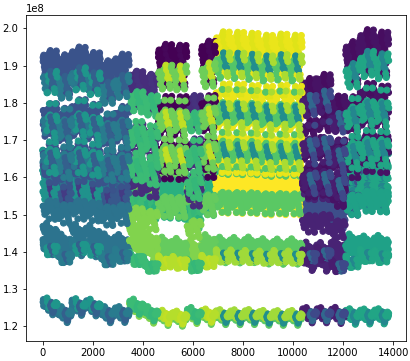} \\
		(a) & (b) \\
	\end{tabular}
	\end{center}
  	\caption{(a) Self-organising feature map (SOFM) with $8 \times 8$ topology after being fitted to all the geological models in three iterations. The neurons are coloured by the OIP corresponding to the model represented by the specific neuron, with darker colours corresponding to a higher associated OIP. (b) visualises the clusters identified by SOFM on the models genomes, with respect to the OIP associated to the models: The $x$-axis represents the model identity number across the full ensemble, while the $y$-axis shows the models' OIP. Models are colour-coded by the cluster they belong to.}
  \label{fig:som}
\end{figure}

Figure~\ref{fig:som}~(a) illustrates our \emph{8 by 8} SOFM fitted to the full model ensemble in three iterations. The map has 64 neurons (the square cells), which correspond to the most representative pseudo-models\footnote{These models have the right structure to be interpreted as evolution of geological properties across an oilfield, but were obtained by averaging geological trends observed in the subsurface rather than derived from physical measurements. As such whether these models represent a plausible geological scenario is yet to be determined.} of the studied oilfield. The neurons are coloured by the amount of OIP associated to the geology represented by each neuron. Given a good grouping of models, the colour should gradually change across the map, with neurons of similar colour located close to each other. 

It can be seen in Figure~\ref{fig:som}~(a) that whilst there are some groups of models with similar OIP values, represented by similarly coloured neurons close together, not all neurons with similar OIP are in the same location, for instance, the three dark bluey-purple areas which are separated quite significantly. Moreover, models with different OIP values are sometimes placed next to each other, e.g., the yellow neuron (low OIP) in the lowest row which is located right next to a dark purple (high OIP) neuron above it. As such we conclude that the map's organisation is only weakly correlated to the OIP.

Figure~\ref{fig:som}~(b) visualises the clusters identified by the SOFM approach. Similarly to the DBSCAN clusters, these do not match the desired pattern of Figure~\ref{fig:clusters}~(a). However, it is interesting to note that the pattern obtained using SOFM differs from that provided by the DBSCAN clusters in Figure~\ref{fig:clusters}~(b), which demonstrates that the two clustering approaches are in-fact driven by different criteria.


\section{Metric for the similarity of target properties}
\label{sec:ml}

The clustering experiments of Section \ref{sec:modelreduction} demonstrate that similarity in a geometrical sense, as defined by Euclidean distance on the models' definitions (the genomes), does not lead to clusters of models with similar behaviour with respect to the target property, OIP. Therefore, the applied clustering algorithms were not able to sort the models into the desired groups. As such, we believe that in order to cluster properly, a similarity metric is required that relates the genome to the associated OIP value as closely as possible. 

We have experimented with several different, user-defined metrics in DBSCAN, but these did not lead to good results and were computationally infeasible. Therefore we explored the approximating the genome-OIP relationship using machine learning techniques. When following this approach a key question is how much labelled data is required to train a regression model sufficiently so it is able predict OIP from the model genome to an acceptable level of accuracy. Both an artificial neural network (NN), and gradient boosting regressor (GB) have been tested, with the objective to understand which approach is most appropriate for this task. We chose a NN because this popular and widely successful technique has proven highly capable in approximating any function with arbitrary precision by a Universal approximation theorem. The GB based approach was also included as this method has been demonstrated to work well with oil and gas datasets \cite{swoop} in other work.  

The neural network is a multilayer perceptron (MLP) with a hidden layer of 30 neurons, \emph{relu} activation function and \emph{adam} solver. The gradient boosting regressor performs 100 boosting stages, with a maximum tree depth of 80 nodes and the \emph{huber} loss function. The Scikit-learn library \cite{pedregosa2011scikit} has been used for both these approaches. When training these machine learning models, we started with 80\% of the geological models for training, and the remaining 20\% for testing. Whilst this showed promising results, with both the NN and GB approaches being able to learn a relationship between the genomes and OIP well enough to satisfy the geologists (predictions within an average range of 1\% of the actual OIP value), this approach requires a full evaluation of 80\% of the models and as such is not practical for real-world use. However it was an important step to validate that ML can predict the Oil In Place (OIP) from the geological models accurately enough to be within bounds acceptable to the geologists.


\begin{figure}[htb]
\centering
	\includegraphics[scale=.45]{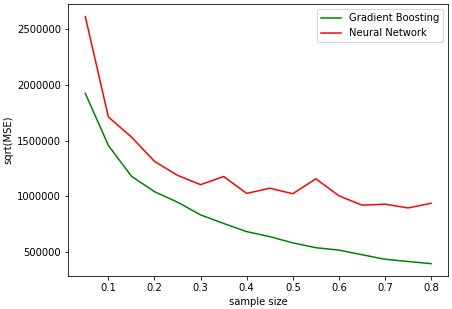}
	\caption{Evolution of average MSE with increasing training data.}
	\label{fig:sample}
\end{figure}

Based on these results we then explored how the size of the training data (the number of models that the NN and GB are trained with) impacts the quality of OIP prediction. Both the NN and GB were trained with samples ranging from 5\% to 80\% of the geological models, in 5\% increments, to understand their sensitivity. The obvious approach would be to simply slice the sample size of models and base our training on this, however these random samples could vary significantly in their coverage of the overall input space. This would significantly impact the OIP prediction and any poor predictions could be more a result of the inadequate splitting of model data, rather than due to a reduced geological model set size. Therefore we carried out the analysis over several different samples, each with independent geological properties, of the same size. This allowed us to obtain a more robust idea about the impact of the sample size and variability of ML model precision, without being impacted by the ML model being trained on very different features.

Figure \ref{fig:sample} illustrates how the square root of the average MSE across the samples of the same size gradually decreases as the number of geological models used to train the ML model increases. It was agreed by domain experts that, for both the NN and GB, training on a sample of 15\% of the geological models produces sufficiently good results. This corresponds to evaluating 2073 of the geological models explicitly to produce acceptable OIP values, reducing the required computation considerably.



\begin{figure}[htb]
    \begin{tabular}{c c}
        \includegraphics[scale=.34]{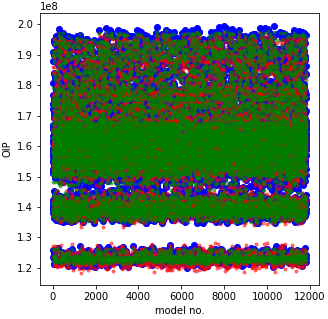} &  \includegraphics[scale=.34]{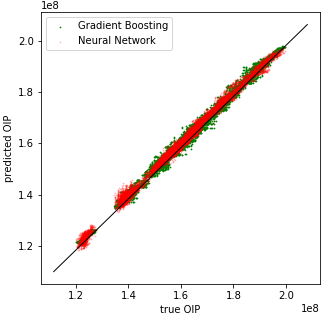} \\
       (a) & (b) \\ 
    \end{tabular}
    \caption{(a) Predictions of OIP by Neural Network (red) and Gradient Boosting (green) models compared to the true OIP (blue) for each geological model in the test set, and (b) a scatter plot comparing the true (observed) and predicted values of OIP across testing all geological models in the test set.} 
    \label{fig:pred-2000}
\end{figure}



To explore this further we trained our NN and GB machine learning models on samples of 2000 models, using the remaining 11824 geological models as the test set.
The plot in Figure~\ref{fig:pred-2000}~(a) illustrates the OIP (on the $y$-axis) corresponding to testing with each of these 11824 geological models, represented by their sequential number on the $x$-axis. The true values of OIP are plotted in blue, and the values predicted by the NN model and GB model in red and green respectively. Figure~\ref{fig:pred-2000}~(b) provides an alternate view, comparing the predicted and true values of OIP across all 11824 geological models in the test set. Based on Figures \ref{fig:pred-2000}~(a) and~(b) it can be seen that, generally speaking, the prediction is fairly close to the true value but that our ML models tend to under predict higher values of OIP. The scatter plot also demonstrates that the error of the Gradient Boosting seems to have a greater amplitude. This is interesting because it can be seen in Figure \ref{fig:sample} that the MSE of GB is lower than that of the NN, so in-fact most predictions will tend to be closer to the truth than the NN, but the outliers are further away. 


\section{Semi-supervised Reduction of the Model Ensemble}
\label{sec:ss-red}

In Section~\ref{sec:modelreduction} we explored the reduction of model ensemble size based on clustering techniques and concluded that the choice of metric function is crucial for generating high quality clusters. Section~\ref{sec:ml} demonstrated how to approximate a useful metric using supervised machine learning. Therefore, in this section, we look to combine these approaches, the SOFM unsupervised clustering technique of Section~\ref{sec:modelreduction} with the supervised learning metric of Section~\ref{sec:ml}. 

Our proposed ensemble reduction follows a semi-supervised approach, where we use a GB regressor, which was found to be optimal in Section \ref{sec:ml}, fitted on a small, random sample of geological models. The intention is that this will estimate the OIP corresponding to the models, which is unknown at this stage in the workflow for real-world data, and we can then use the difference of predicted OIP values to define the similarity between two models. Whilst the initial selection of models to be used with the GB regressor will likely be far from optimal, in terms of capturing all the geological features, it was our hypothesis that as this is only used as a basis for the similarity metric, it could still be sufficient. 

\begin{figure}[htb]
	\begin{center}
    \begin{tabular}{c c}
        \includegraphics[scale=.161]{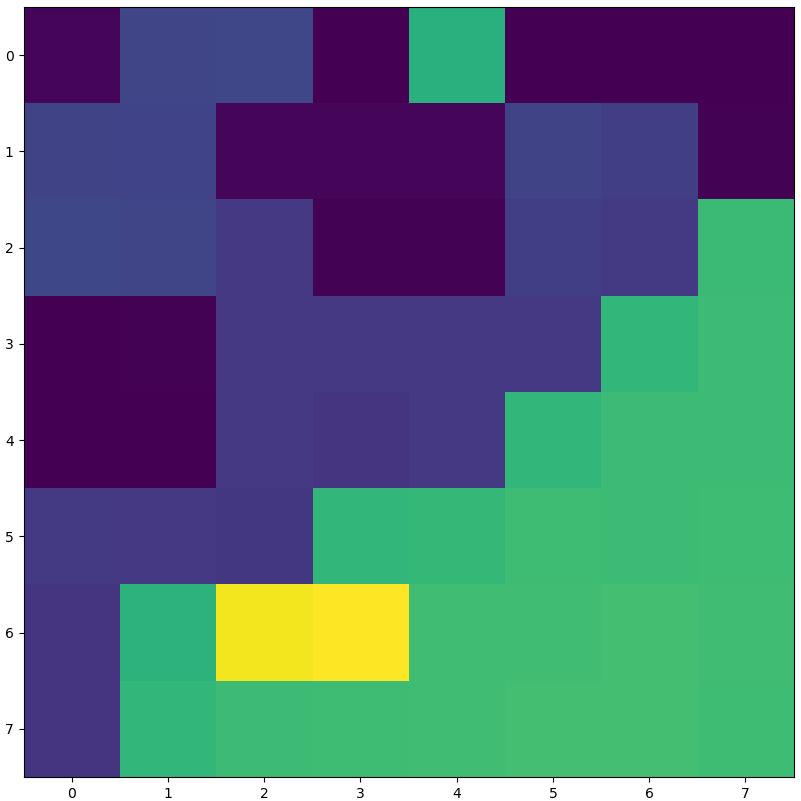} & \includegraphics[scale=.27]{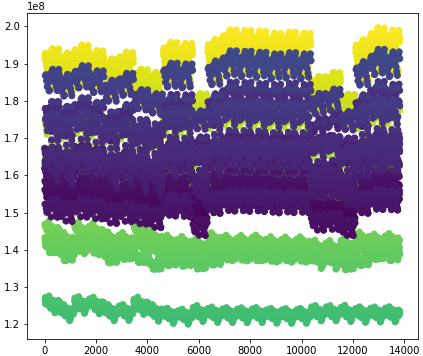} \\
       (a) & (b) \\ 
    \end{tabular}
   	\end{center}
    \caption{(a) An $8 \times 8$ SOFM fitted to all the genomes in three iterations using  values of OIP predicted by a GB regressor as the metric function. (b) A visualisation of the clusters identified by the SOFM, where every cluster is defined as a set of models that map to the same neuron in the fitted SOFM. Note that the models are colour-coded by the cluster they belong to and not by their associated OIP.} 
    \label{fig:som_gb}
\end{figure}

This metric is then used by SOFM, organising the geological models based on the similarity of their expected geological behaviour. This organisation is driven by the OIP prediction and identifies the representative pseudo-models. The results from this semi-supervised approach, the fitted SOFM and the corresponding clusters, are illustrated in Figure~\ref{fig:som_gb}, and it can be seen that the pattern closely approximates the ideal clustering of Figure~\ref{fig:clusters}~(a). This means that only 64 individual models are required, one from each cluster generated by the SOFM and less than 0.5\% of the overall model set size, to fully represent the oilfield. 

It is worth noting that once we have the GB model fitted, one can simply build a histogram of the predicted OIP values and use the histogram bins as clusters. However, this approach might not provide a good reduction of the ensemble, because it is not clear which model should be chosen to represent each bin. This is because, as discussed in Section~\ref{sec:modelreduction}, the link between geological behaviour and a models' genome is not a simple one. Consequently, peterophyscists could mistakenly choose very similar genomes to represent considerably different geological behaviours. The SOFM provides pseudo-models which represent the features that are driving OIP values, and that ultimately the geologists might wish to use to study the possible scenarios representing observations in the oilfield.
%
%


\section{Conclusions and further work}
\label{sec:conc}
\noindent Motivated by a real-world problem, in this paper we have explored how to most effectively reduce an ensemble of geological models down to a subset that truly reflects the variety of the original ensemble. This is required because the evaluation of the full ensemble is computationally expensive and infeasible for the ever increasing sizes of data being generated. We proposed a two-step approach, significantly reducing the size of the geological model ensemble whilst maintaining acceptable accuracy, based on evaluating a random 15\% sample of the geological models. 

In the first step, we used unsupervised learning techniques to group the geological models.
The genomes within a cluster vary considerably, as demonstrated  by the results of clustering using Euclidean distance as the metric. We have found out that the most suitable metric function for the unsupervised techniques is, in this case, based on the value of OIP corresponding to each model, however, this value is not known until the geological models are fully evaluated, which we are trying to avoid. Therefore, we approximate the metric using supervised machine learning techniques.

We found that a Gradient Boosting Regressor is optimal in predicting the target property, Oil in Place (OIP), and this machine learning model fitted to only 2000 models (less than 15\% of the ensemble) predicts the OIP without any systematic bias with mean absolute error (MAE) of $2.4 \times 10^5$ on values ranging from $1.2 \times 10^8$ to $2 \times 10^8$, which is less than 1\% of the value on average. 
%


We have found that self-organising feature maps are most suitable for clustering in combination with a bespoke metric, which is based on a rough estimate of OIP from a regression model on a random sample of the geological models. Using this approach we have demonstrated that our method can identify 64 models, less than 0.5\% of the overall geological model set, which are suitably representative of the entire state space. 

Work is on-going, and an area for further exploration is in the selection of models for training the regression in the semi-supervised clustering approach. Going from a random sample to a more informed approach could improve our accuracy further, as would hyper-parameter tuning. We currently working with the petroleum industry to apply our approach to other oilfields and locations with more complex definitions. A key question is whether the insights and lessons we learnt will transfer directly, or whether further work is needed in other locations. 

\pagebreak




\bibliographystyle{./bibliography/IEEEtran}
\bibliography{./bibliography/IEEEexample}

\end{document}